\documentclass{article}
\usepackage{times}

\usepackage[final]{corl_2020}
\usepackage{color}
\usepackage{multicol}
\usepackage{rigidmath}
\usepackage{stfloats}
\usepackage{siunitx}
\usepackage{graphicx}
\usepackage[font=small]{caption}
\usepackage{subcaption}
\usepackage{amsfonts}
\usepackage{url}

\title{ContactNets: Learning Discontinuous Contact Dynamics with Smooth, Implicit Representations}

\author{Samuel Pfrommer\thanks{The first two authors contributed equally to this work.} , Mathew Halm$^*$, and Michael Posa \\
GRASP Laboratory, University of Pennsylvania\\
\texttt{\{spfrom, mhalm, posa\}@seas.upenn.edu}}

\begin{document}



%

\maketitle

\begin{abstract}
Common methods for learning robot dynamics assume motion is continuous, causing unrealistic model predictions for systems undergoing discontinuous impact and stiction behavior.
In this work, we resolve this conflict with a smooth, implicit encoding of the structure inherent to contact-induced discontinuities.
Our method, \textit{ContactNets}, learns parameterizations of inter-body signed distance and contact-frame Jacobians, a representation that is compatible with many simulation, control, and planning environments for robotics.
We furthermore circumvent the need to differentiate through stiff or non-smooth dynamics with a novel loss function inspired by the principles of complementarity and maximum dissipation. 
Our method can predict realistic impact, non-penetration, and stiction when trained on 60 seconds of real-world data.
\end{abstract}


\section{Introduction}

To effectively perform a wide variety of manipulation tasks, intelligent robots must understand not only how they can affect the motion of objects in their environment, but also how objects in their environment interact with one another.
While recent accomplishments in planning \cite{Mordatch2012,Posa2013a} and control \cite{Aydinoglu2020}, suggest that a model of the robot-environment system's dynamics is a highly useful formalism for capturing this behavior, producing an accurate model from data is a challenging task for manipulation systems due to the complex behaviors induced by frictional contact.

\begin{figure}[h]
\centering
\subcaptionbox{1D System\label{subfig:1DPoint}}{\includegraphics[width=0.15\textwidth]{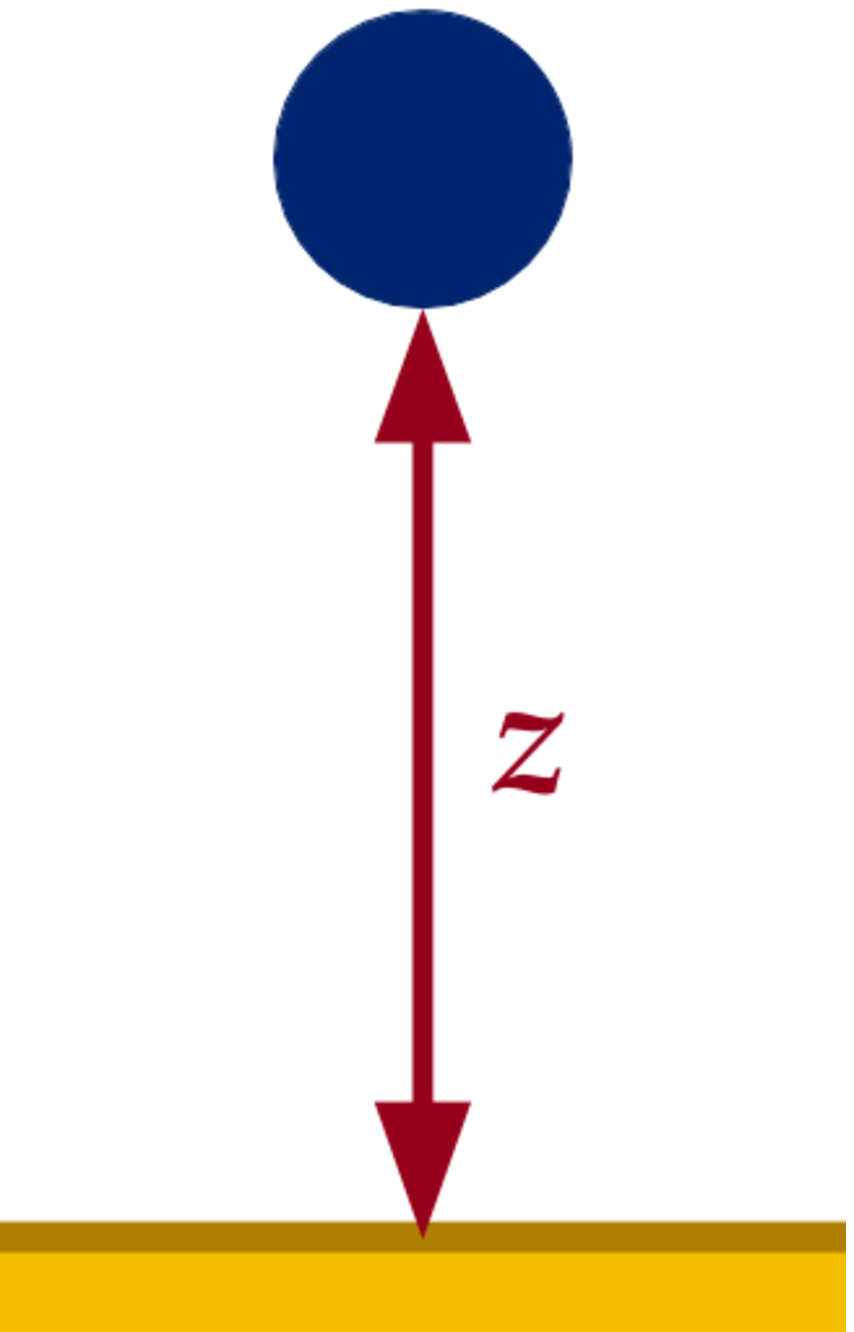}
\vspace{4ex}
}
\subcaptionbox{Model Predictions\label{subfig:1DPrediction}}{\includegraphics[width=0.54\textwidth]{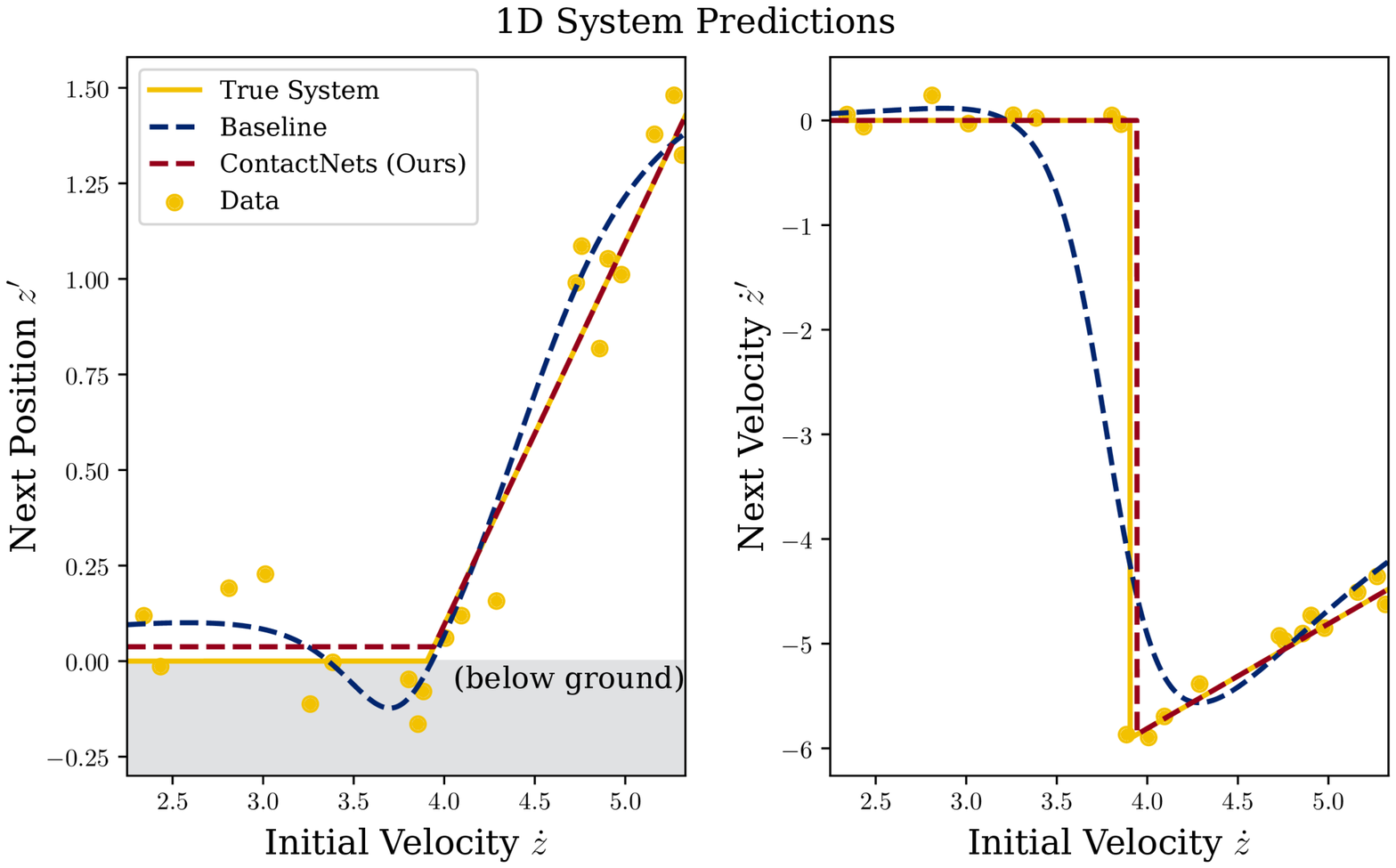}}
\subcaptionbox{Loss Landscape\label{subfig:1DLoss}}{\includegraphics[width=0.27\textwidth]{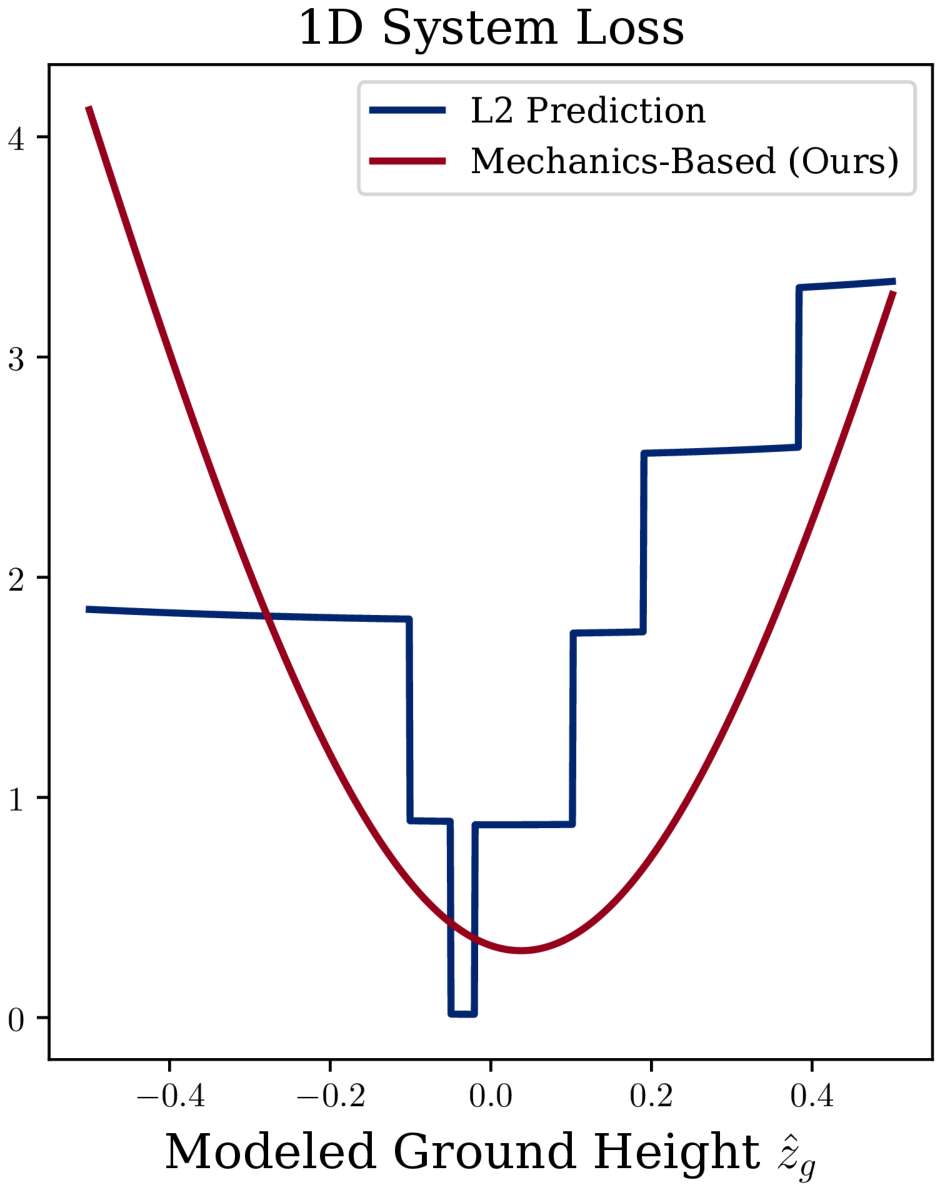}}

\caption{(\subref{subfig:1DPoint}) A 1-D point mass system illustrates contact's pathological nature in model learning. (\subref{subfig:1DPrediction}) A projection (solid yellow) and noisy sampling (yellow dots) of the system's equations of motion onto the $z = 1$ plane is shown. Fitting a DNN (blue) directly to the data is unable to capture the velocity discontinuity well, and predicts significant ground penetration. Our method (red) instead uses gravitational acceleration as a prior, and learns the height of the contact surface $\hat z_g$; though noise prevents a perfect fit, the resulting dynamics qualitatively and quantitatively exceed the unstructured approach. (\subref{subfig:1DLoss}) Supervised learning of $\hat z_g$ with $L_2$ loss and gradient descent is ill-posed, as discontinuities in the model are propagated to the loss landscape. We instead ensure reliable training via a smooth novel loss (red) based on contact mechanics. A detailed explanation of this example can be found in Appendix \ref{subsec:1DAppendix}. \label{fig:1DExample}
}
\vspace{-.5cm}
\end{figure}

Many methods for learning a dynamical system attempt to fit a universal function approximator to the system's equations of motion \cite{Chua2018,Deisenroth2011,Bauza2017probabilistic,Fazeli2017} or inverse dynamics \cite{Meier2016}, yet often the underlying inductive biases conflict with the the nature of frictional contact. Two ubiquitous representations, fully-connected deep neural networks (DNNs) and Gaussian processes regression with squared-exponential kernels \cite{Deisenroth2011,Bauza2017probabilistic,Fazeli2017}, are biased towards similar interpretations of Occam's razor: the best parameterization (i.e. simplest explanation) is the smoothest interpolator \cite{Belkin2019} or an infinitely-differentiable regressor \cite{Rasmussen2005} of the data.
However, physics-based analysis predicts discontinuity \cite{Coumans2015}, non-uniqueness \cite{Halm2019}, and/or extreme curvature \cite{Todorov2014} within the equations of motion for systems undergoing frictional contact, and furthermore areas of state-input space crucial to locomotion and manipulation (e.g. footfalls and grasping) are \textit{precisely} where these irregularities arise. This conflict manifests as poor model predictions even in simple scenarios, as illustrated in Figure \ref{fig:1DExample}.

Some works attempt to capture discontinuity with multi-modality \cite{Fazeli2019}.
However, unstructured multi-modality is computationally intractable for multi-contact behaviors as the number of modes is extremely large, even in toy systems \cite{Huang2020}.
DNNs can alternatively be conditioned to generate contact-like behaviors by embedding a differentiable physics simulator directly into their structure \cite{Battaglia2016,BelbutePeres2018,Li2018}.
However, differentiating through simulation is numerically challenging if discontinuity is approximated with high-curvature \cite{Kolev2015}, and results from these methods have been limited to simulation and quasi-static real-world interaction with highly-compliant objects.
Embedding tactile and force/torque sensors into the robot and/or environment can be effective for learning robot-objects interaction \cite{Calandra2015,Fazeli2019}, though using such methods to learn object-object interactions would necessitate embedding countless sensors in the robot's environment.

We present a novel approach to the problem of learning frictional contact behaviors, ContactNets, which eliminates these pervasive difficulties and leads to a well-conditioned problem that is amenable to data-efficient learning.
The main contribution of this paper is a reparameterization of the learning problem that effectively uses the inductive bias of DNNs for frictional contact without requiring \textit{any} contact or force sensing.
Inspired by both frictional contact mechanics \cite{Anitescu1997,Stewart1996a,Todorov2014,Coumans2015} and implicit representations in deep learning \cite{Park2019}, we implicitly parameterize and learn discontinuous contact behaviors as continuous inter-body signed distance functions and contact-frame Jacobians.
This representation is equivalent to the parameterization of contact internal to simulators for robotics (e.g. \cite{drake,Coumans2015,Anitescu1997,Stewart1996a,Todorov2014}), permitting generation of realistic discontinuous behaviors with efficient numerical optimization.
At training time, we circumvent the numerical challenges of differentiating through discontinuous simulators with a novel loss function inspired by the principles of complementarity and maximum dissipation.
We evaluate our method on a real-world, dynamic, 3D frictional contact scenario and compare its performance to an unstructured baseline.

\section{Background}

A robotic manipulator interacting with rigid objects and environment can be modeled with inputs $\Input$ (e.g. motor torques) and states $\State = [\Configuration; \Velocity]$, where $\Configuration$ represents the robot's configuration and object poses and $\Velocity$ represents velocities. The discrete-time dynamics $\State' = f(\State,\Input)$ of this system can be formulated as a generalization of Newton's second law emerging from Lagrangian mechanics:
\begin{equation}
	\Mass(\Configuration)(\Velocity' - \Velocity) = \NetForce_{net}(\State,\Input)\,.\label{eq:NewtonsSecondLaw}
\end{equation}
$\Mass(\Configuration)$ represents inertial quantities, and $\NetForce_{net}$ is the net generalized impulse over the timestep\footnote{We use the letter $\NetForce$ to denote behaviors emergent from contact forces for notational clarity, but note that $\NetForce_{net}$ is the net \textit{impulse} over the timestep, i.e. $(\textrm{Net Force}) \times \Delta t$.}. Configurations are updated via integration of the velocity\footnote{For 3D systems, $\Configuration$ and $\Velocity$ often use different coordinates (e.g. quaternions and angular velocities) which obey $\GeneralizedVelocityJacobian (\Configuration) \Velocity = \TimeDiff{}{}\Configuration$, where $\GeneralizedVelocityJacobian$ is a Jacobian. \eqref{eq:QDotFromV} and \eqref{eq:NormalComplementarity} become $\Configuration' - \Configuration = \GeneralizedVelocityJacobian(\Configuration)\Velocity'\Delta t$ and $\Jn[i] = \nabla_{\Configuration} \Gap_{n,i}\GeneralizedVelocityJacobian$.}, e.g.
\begin{equation}
	\Configuration' - \Configuration =  \Velocity' \Timestep \label{eq:QDotFromV}\,.
\end{equation}
For a system experiencing up to $\Contacts$ contact interactions, $\NetForce_{net}$ can be decomposed as
\begin{equation}
	\NetForce_{net}(\State,\Input) = \NetForce_{s}(\State,\Input) + \sum_{i=1}^\Contacts \J[i](\Configuration)^T\Force[i] \label{eq:NetImpulseDecomposition}\,.
\end{equation}
$\NetForce_s$ aggregates smooth, non-contact impulses which emerge from potential (e.g. gravitational), gyroscopic, and input impulses; and for each $i$, $\J[i]^T\Force[i]$ is the net impulse due to the $i$th contact. Here, $\J[i]= [\Jn[i]; \Jt[i]]$ is the configuration-dependent contact Jacobian which maps generalized velocities into Euclidean velocities in the $i$th contact frame normal ($\Jn[i]$) and tangential ($\Jt[i]$) directions. $\Force[i] = [\NormalForce[i]; \FrictionForce[i]]$ are the contact-frame normal impulses $\NormalForce[i]$, which resist interpenetration; and frictional impulses $\FrictionForce[i]$, which resist sliding motion between the contacting surfaces.


The underlying mathematics in simulators for rigid robots and environments (e.g. MuJoCo \cite{Todorov2014}, Bullet \cite{Coumans2015}, Drake \cite{drake}, and others \cite{Anitescu1997, Stewart1996a}) stray very little from the structure in \eqref{eq:NewtonsSecondLaw}--\eqref{eq:NetImpulseDecomposition}, and are primarily differentiated in their methodology for calculating the contact impulses $\Force[i]$. However, many models approximate the same two essential characterizations of contact behavior:
\begin{itemize}
	\item \textbf{Normal complementarity}: The signed distance function $\Gap_n (\Configuration) = [\Gap_{n,1};\dots;\Gap_{n,m}] \in \Real^\Contacts$ captures contact geometry as inter-body distances.
Because bodies cannot interpenetrate and normal forces only push bodies apart when they touch, for each contact $i$,
\begin{align}\label{eq:NormalComplementarity}
	\Jn[i] = \nabla_{\Configuration} {\Gap_{n,i}}\,, && \Gap_{n,i} \geq 0\,, && \NormalForce[i] \geq 0\,, && \Gap_{n,i}\NormalForce[i] = 0\,.
\end{align}
\item \textbf{Maximal dissipation}: Many friction models pick $\FrictionForce[i]$ from an admissible set $\FrictionForceSet[i]$ such that mechanical power loss is maximized \cite{Stewart1996a,Zhou2016}:
\begin{equation}\label{eq:MaximumDissipation}
	 \FrictionForce[i] \in \underset{\FrictionForce[i]' \in \FrictionForceSet[i]}{\arg\min} \quad \FrictionForce[i]' \cdot \Jt[i](\Configuration)\Velocity\,,
\end{equation}
For instance, Coulomb's friction model with coefficient of friction $\FrictionCoeff[i]$ uses
\begin{align}
	\FrictionForceSet[i] = \Braces{\FrictionForce[i]: \TwoNorm{\FrictionForce[i]} \leq \FrictionCoeff[i]\NormalForce[i] }\,.\label{eq:CoulombCone}
\end{align}
For nonzero velocities (sliding),  \eqref{eq:MaximumDissipation} has the closed form solution
\begin{equation}
	 \FrictionForce[i] = -\frac{\Jt[i]\Velocity}{\TwoNorm{\Jt[i]\Velocity}}\FrictionCoeff[i]\NormalForce[i]\,.\label{eq:ExplicitMaximumDissipation}
\end{equation}
\end{itemize}

\section{Related work}
A large body of research (\cite{Mordatch2012,Posa2013a,Fazeli2019,Halm2018,Huang2020,BelbutePeres2018,Li2018,Calandra2015} and others) in learning and robotics has recognized many challenges in identifying and controlling frictional contact behaviors. Even smooth contact-induced motion is complex due to partial observability, multi-modality, and stochasticity. Several recent works focus on objects sliding on flat surfaces via pushing; \citet{Zhou2016} learn a set-valued representation of frictional forces via convex optimization and \citet{Bauza2017probabilistic} learn a pusher-slider system's dynamics via Gaussian processes regression. \citet{Fazeli2017} learn the mapping from pre-impact velocity to post-impact velocity of a planar object falling onto a flat surface, and show superior performance to mechanics-based models. \citet{Ajay2018} as well as \cite{Fazeli2017} utilize \textit{residual physics}, in which the gap between a physics simulator and real system's motion is learned. While these methods produce rich descriptions of frictional contact, they all critically assume \textit{a priori} knowledge of what contacts are active. In this work we instead focus on the separate challenge of learning \textit{where and when} discontinuous and non-smooth behaviors including impact and stick-slip transitions occur.

Many methods tackle this problem by inserting a mechanics-based or learned physics model with analytical gradients in an end-to-end optimization framework, such as a DNN.
\citet{BelbutePeres2018}, for instance, develop a piecewise-continuous, mechanics-based simulator, and differentiate through continuous motion assuming known, fixed, discontinuous impacts. 
Ignoring the discontinuities can adversely affect the prediction loss landscape (see Figure \ref{subfig:1DLoss}), and makes learning contact geometries impossible as no gradients are propagated back to inter-body distance.
Other methods represent all contact behaviors as a learned, fully differentiable DNN.
\citet{Battaglia2016} structure multi-object simulation as pairwise interactions, but even a single object experiencing impact is difficult to model as a DNN (See Sections \ref{sec:Procedure}--\ref{sec:Results}), and their method is not tested on real-world data.
\citet{Li2018} extend this method to real-world deformable body manipulation, but the extreme compliance and quasi-static motion present in their setting does not exhibit the discontinuities that are fundamental to essential robotics tasks.
Other works model discontinuity as multi-modality; \citet{Fazeli2019} for instance learn a hierarchical model that does not embed strong physics-based priors, yet is capable of segmenting a handful of distinct contact modes when pushing Jenga blocks out of a tower. However, such methods typically scale in complexity with the number of smooth modes, which grows combinatorially with the number of objects in the environment.
\citet{Calandra2015} instead leverage the continuity of the state-update equations \eqref{eq:NewtonsSecondLaw}, \eqref{eq:NetImpulseDecomposition} in the contact forces $\Force[i]$, and learn the mapping from current state \textit{and} sensed contact forces to next state. This approach requires full observation of $\Force[i]$, which makes multi-step prediction impossible and furthermore requires sensors to be embedded in the environment for multi-object manipulation. 

\citet{Fazeli2017Estimation} is perhaps the closest in spirit to our method, in which a nonlinear optimization problem (NLP) is developed to identify a handful of parameters for planar systems with contact, and furthermore establishes a complementarity-based loss similar to ours. However, computation of the NLP grows intractably with the number of contacts and datapoints. Our bilevel optimization instead uses efficient convex optimization in the inner loop, and scalable, gradient-based unconstrained optimization in the outer loop, enabling identification of complex geometries from a large dataset.

\section{Approach}\label{sec:Approach}

\begin{figure}
	\includegraphics[width=\textwidth]{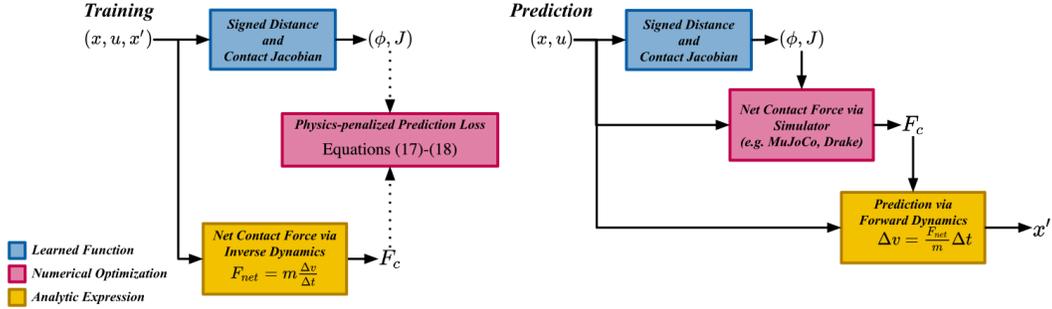}
	\centering
	\caption{Computation graphs for the proposed model. At training time (left), contact-related terms $(\Gap,\J)$ are generated from a state transition, and a loss on the model is computed as the solution to an optimization problem that quantifies how well the the terms fit the transition. Prediction is performed by inputting the learned $(\Gap,\J)$ to a contact simulation environment, which typically solves a conic program (e.g MuJoCo \cite{Todorov2014}), unconstrained nonlinear program (e.g. Drake \cite{drake}), or linear complementarity program (e.g. Bullet \cite{Coumans2015} and others \cite{Stewart1996a,Anitescu1997}) to generate feasible contact forces.\label{fig:TrainingandPrediction}}
\vspace{-0.5cm}
\end{figure}
We now present \textit{ContactNets}, our approach to learning a model explicitly capable of predicting discontinuous contact impulses. We consider systems containing rigid robots and environments, so that the model class \eqref{eq:NewtonsSecondLaw}--\eqref{eq:NetImpulseDecomposition} readily applies.
For such systems, identification of the contact-free dynamics, parameterized by inertial quantities $\Mass$ and non-contact impulses $\NetForce_{s}$, has been studied by roboticists for decades, and many algorithms exist for learning accurate models (e.g. in \cite{Khosla1985,Traversaro2013}).
Therefore, we additionally assume that $\Mass$ and $\NetForce_{s}$ are known (either from a hand-designed or learned model), and focus on learning to predict the contact impulses, as in \citet{Calandra2015}.

An overview of our approach is given in Figure \ref{fig:TrainingandPrediction}, which depicts how \textit{ContactNets} models are trained and tested.
All contact behaviors in \eqref{eq:NewtonsSecondLaw}--\eqref{eq:NetImpulseDecomposition} are determined solely by the inter-body distance $\Gap_{n,i}(\Configuration)$, contact Jacobian $\J[i](\Configuration)$, and friction coefficient $\FrictionCoeff[i]$.
Furthermore, while contact dynamics can be complex, discontinuous, and multimodal, $\Gap_{n,i}$ and $\J[i]$ are often modeled as simple, smooth functions.
Therefore, at training time we learn functional approximations of these quantities using only state transitions $\mathcal D = (\State_j,\Input_j,\State_j')_{j\in 1,\dots,D}$.
This process does not require \textit{any} tactile or force/torque sensor information, which crucially does not require instrumenting each object with contact sensing hardware, an impractical requirement outside of a tightly-controlled research environment.
This is in contrast to other methods (e.g. \cite{Calandra2015,Fazeli2019}) which learn a mapping from the outputs of contact force sensing hardware to $\NetForce_{net}$ directly.
At test time, we can predict the net impulse $\NetForce_{net}$ using the same mathematics from well-established simulation techniques, including MuJoCo \cite{Todorov2014}, Bullet \cite{Coumans2015}, Drake \cite{drake}, and others \cite{Anitescu1997, Stewart1996a}.
Furthermore, for generating intelligent behaviors, the learned parameterization is compatible with several dynamic planning algorithms \cite{Mordatch2012,Posa2013a} and controllers \cite{Aydinoglu2020} that have specifically been designed to handle the challenges of frictional contact.

\subsection{Model parameterization}\label{subsec:ModelParameterization}
We construct an approximation of inter-body distances $\Gap_n^{\Parameters}(\Configuration)$ with parameters $\Parameters$, and we calculate $\Jn[i]^{\Parameters}(\Configuration) = \nabla_{\Configuration}\, \Gap_{n,i}^{\Parameters}$ using back-propagation.
$\Gap_n^{\Parameters}$ can embed strong geometric priors (e.g. a polytope contacting flat ground), enabling learning of a handful of features from sparse training data.
Alternatively, essentially arbitrary object and robot geometries can be represented if $\Gap_n^{\Parameters}$ is a DNN.
For some systems (e.g. a polytope or strictly convex shape contacting flat ground), the tangential contact Jacobian $\Jt[i]$ is also the gradient of some function of configuration $\Configuration$, so we create another function approximator $\Gap_t^{\Parameters}$ and calculate $\Jt[i]^{\Parameters}(\Configuration) = \nabla_{\Configuration} \Gap_{t,i}^{\Parameters}$.

In order to make our formulation tractable and well-posed, we also assume the following:
\begin{itemize}
	\item 
\textbf{Configuration-dependent Coulomb friction:} To simplify the maximal dissipation constraint \eqref{eq:MaximumDissipation}, we assume that, given a particular contact location, friction behaves according to Coulomb's model--i.e. for some configuration-dependent friction coefficient $\FrictionCoeff[i](\Configuration)$, \eqref{eq:CoulombCone} and \eqref{eq:ExplicitMaximumDissipation} hold. Due to scale invariance in the net force calculation, we equivalently assume $\FrictionCoeff[i](\Configuration) = 1$ and learn the lumped term $\Jt[i]^{\Parameters}(\Configuration) = \FrictionCoeff[i](\Configuration)\Jt[i](\Configuration)$.
\item \textbf{Discrete-time contact activation:} The complementarity condition \eqref{eq:NormalComplementarity} ideally holds for all times, but the dataset only contains a discrete-time sampling of the state.
We therefore make the approximating assumption that if the $i$th contact is active at all during a time-step, then it is still active at the \textit{end} of the step (a standard approach in simulation \cite{Stewart1996a}). That is,
\begin{align}\label{eq:ApproximateNormalComplementarity}
	\Gap_{n,i}(\Configuration') \geq \ZeroVector\,, && \NormalForce[i] \geq \ZeroVector\,, && \Gap_{n,i}(\Configuration')\NormalForce[i] = 0\,.
\end{align}
 \eqref{eq:ApproximateNormalComplementarity} holds approximately for inelastic impacts or high sampling frequencies.
\end{itemize}

 While the contact forces in real systems do not behave exactly according to these assumptions, models in this class are accurate enough to produce agile and accurate motion in locomotion and manipulation tasks (e.g. \citet{Fallon15}).
 
\subsection{Loss formulation}

To train our models, we need a loss function $\Loss(\Parameters,\State,\Input,\State')$ that captures how well they explain a particular transition $(\State,\Input,\State')$. We observe that, even in the absense of sensing for the \textit{individual} contact impulses $\Force[i]$, knowledge of the contact-free dynamics enables a ground-truth observation of the \text{net} contact impulse $\NetForce_{c}$ from the data:
\begin{equation}
    \NetForce_{c,data}(\State,\Input,\State') =
    \Mass(\Configuration)(\Velocity'-\Velocity) - \NetForce_s(\State,\Input).\label{eq:ForceFromData}    
\end{equation}
A key insight, taking inspiration from multi-contact simulation \cite{Stewart1996a} and planning \cite{Posa2013a}, is to hypothesize a candidate set of contact impulses $\Force$.
Given such a $\Force$, it is straightforward to capture how well the model explains $(\State,\Input,\State',\Force)$ by a) determining how realistic $\Force$ is, quantifying violation of complementarity and maximum dissipation, and b) calculating how closely the force $\Force$ matches $\NetForce_{c,data}$. We establish the following costs on $(\Parameters,\State,\Input,\State',\Force)$: 
\begin{itemize}
	\item \textbf{Prediction quality:}
$\Force$ should explain the observed contact forces $\NetForce_{c,data}(\State,\Input,\State')$:
\begin{equation}\label{eq:PredictionCost}
	l_1(\Parameters,\State,\Input,\State',\Force) = \Bigg\lVert {\sum_i \J[i]^{\Parameters}(\Configuration)^T\Force[i]-\NetForce_{c,data} }  \Bigg\rVert^2\,.
\end{equation}
While \eqref{eq:PredictionCost} is similar in spirit to $L_2$ loss on the output of a simulator, its mathematical behavior is fundamentally different due to its dependence on the unknown $\Force$.
\item \textbf{Contact activation:}
Forces should only be applied when contact is established ($\Gap_{n,i}(\Configuration')\NormalForce[i] = 0$):
	\begin{equation}\label{eq:ActivationCost}
	l_2(\Parameters,\State,\Input,\State',\Force) = \sum_i \Gap_{n,i}^{\Parameters}(\Configuration')^2\TwoNorm{\Force[i]}^2\,.
	\end{equation}
\item \textbf{Non-penetration:} The motion generated by $\Force[i]$ should not cause penetration. Manipulating \eqref{eq:NewtonsSecondLaw}--\eqref{eq:QDotFromV}, we can estimate how $\Force$ would affect $\Gap_{n,i}^{\Parameters}$ over the time-step as
\begin{align}
	\tilde \Velocity'(\State,\Force) & = \Velocity + \Mass^{-1}(\NetForce_s + \sum_i \J[i]^{\Parameters}(\Configuration)^T\Force[i])\,,\label{eq:VelocityPlusLinear} \\
	\tilde \Gap_{n,i}'(\State,\Force) &= \Gap_i^{\Parameters}(\Configuration) +  \Jn[i]^{\Parameters}(\Configuration) \tilde \Velocity'(\State,\Force)\Delta t\,.\label{eq:GapPlusLinear}
\end{align}
We then penalize negativity of the predicted signed distance $\tilde \Gap_{n,i}'(\State,\Force)$:
\begin{equation}\label{eq:PenetrationCost}
	l_3(\Parameters,\State,\Input,\State',\Force) = \sum_i \min(0, \tilde \Gap_{n,i}'(\State,\Force))^2\,.
	\end{equation}
\item \textbf{Maximal dissipation:} Friction forces $\FrictionForce[i]$ must be chosen such that power loss is maximized. We penalize violation of \eqref{eq:CoulombCone}, scaled by $\TwoNorm{\Jt[i]^{\Parameters}\Velocity'}$:
\begin{equation}\label{eq:DissipationCost}
 	l_4(\Parameters,\State,\Input,\State',\Force) = 
 	\sum_i \TwoNorm{\TwoNorm{\Jt[i]^{\Parameters}(\Configuration)\Velocity'}\FrictionForce[i] + \NormalForce[i]\Jt[i]^{\Parameters}(\Configuration)\Velocity'}^2 \,.
	\end{equation}
\end{itemize}
Finally, we choose a positive loss $\Loss(\Parameters,\State,\Input,\State')$ that is equal to $0$ if a \textit{single}, feasible set of contact forces $\Force$ causes each of the costs $l_k$ to be $0$:
\begin{align}
 	\Loss(\Parameters,\State,\Input,\State') = \underset{\Force}{\min} \quad & \sum_k l_k(\Parameters,\State,\Input,\State',\Force)\,,\label{eq:FinalQPLoss} \\
 	\SubjectTo \quad & \NormalForce[i] \geq 0\,, \TwoNorm{\FrictionForce[i]} \leq \NormalForce[i]\,.\label{eq:QPFrictionConstraint}
\end{align}
Each of the individual costs \eqref{eq:PredictionCost}, \eqref{eq:ActivationCost}, \eqref{eq:PenetrationCost}, \eqref{eq:DissipationCost} are convex piecewise-quadratic in $\Force$. Therefore, with appropriately chosen slack variables, we can pose \eqref{eq:FinalQPLoss} as a tractable, feasible, and convex program, allowing for its gradient to be efficiently computed through sensitivity analysis \cite{Amos2017}.

This structure is unexpected because the converse problem of predicting the transition via simulation is often formulated as a non-convex problem due to the complementarity constraint \eqref{eq:ApproximateNormalComplementarity}, as both the signed distance $\Gap_n(\Configuration')$ and forces $\Force$ must be solved for simultaneously \cite{Stewart1996a}.
It is because the transition is \textit{already observed} and only $\Force$ is unknown that this difficulty is circumvented.

\section{Experimental procedure}\label{sec:Procedure}

\begin{figure*}[ht]
\begin{minipage}{\hsize}
\begin{minipage}{.49\linewidth}
    \includegraphics[width=.65\textwidth]{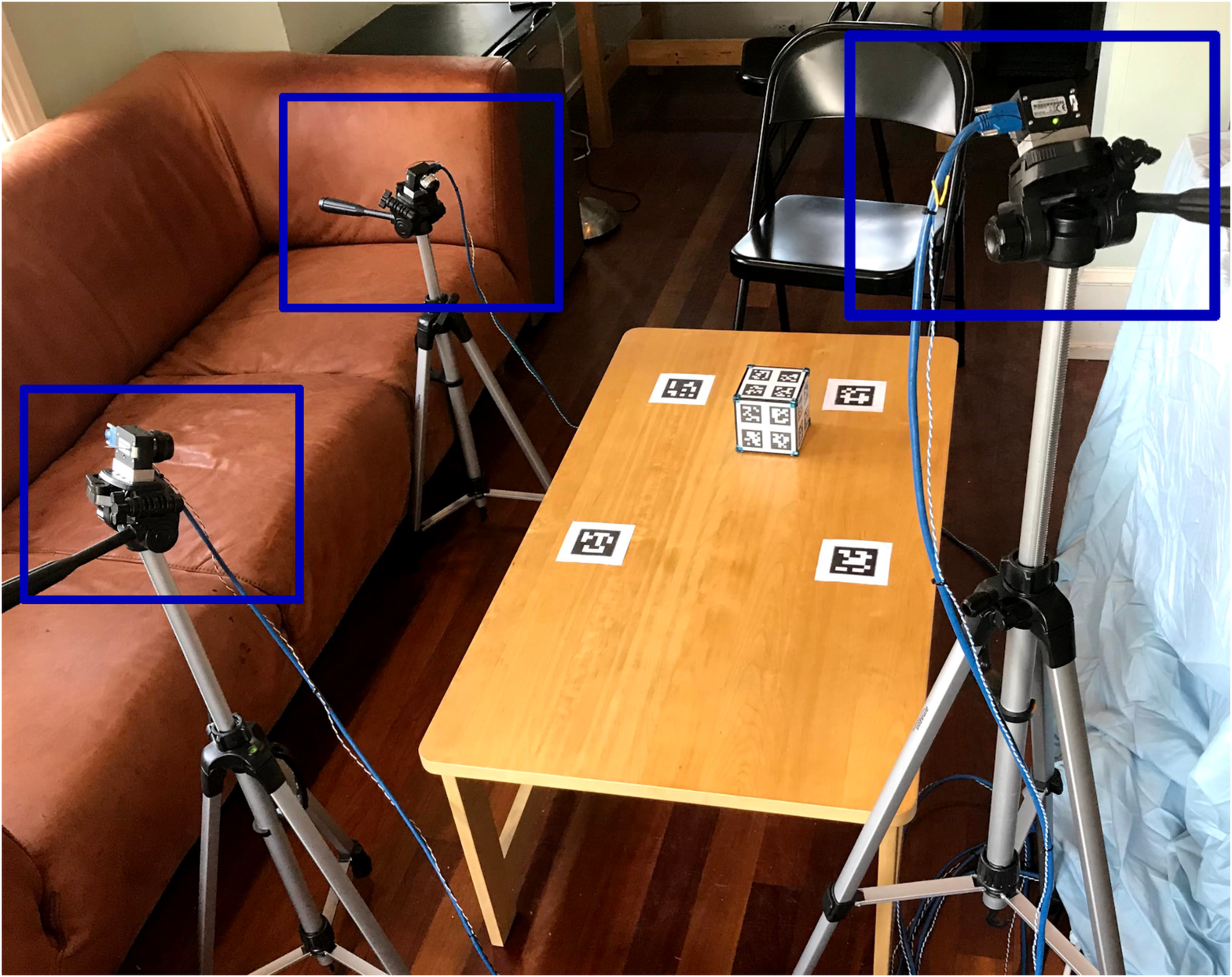}
    \centering
    \caption{\label{fig:setup}The experiment was designed to evaluate complex, three dimensional contact.
    A cube was tossed against the planar ground, producing impacts and both sticking and sliding frictional contact. Cube position and orientation are tracked by three cameras, outlined in blue.}
\end{minipage}
\hspace{0.01\textwidth}
\begin{minipage}{.49\linewidth}
\includegraphics[width=.65\textwidth]{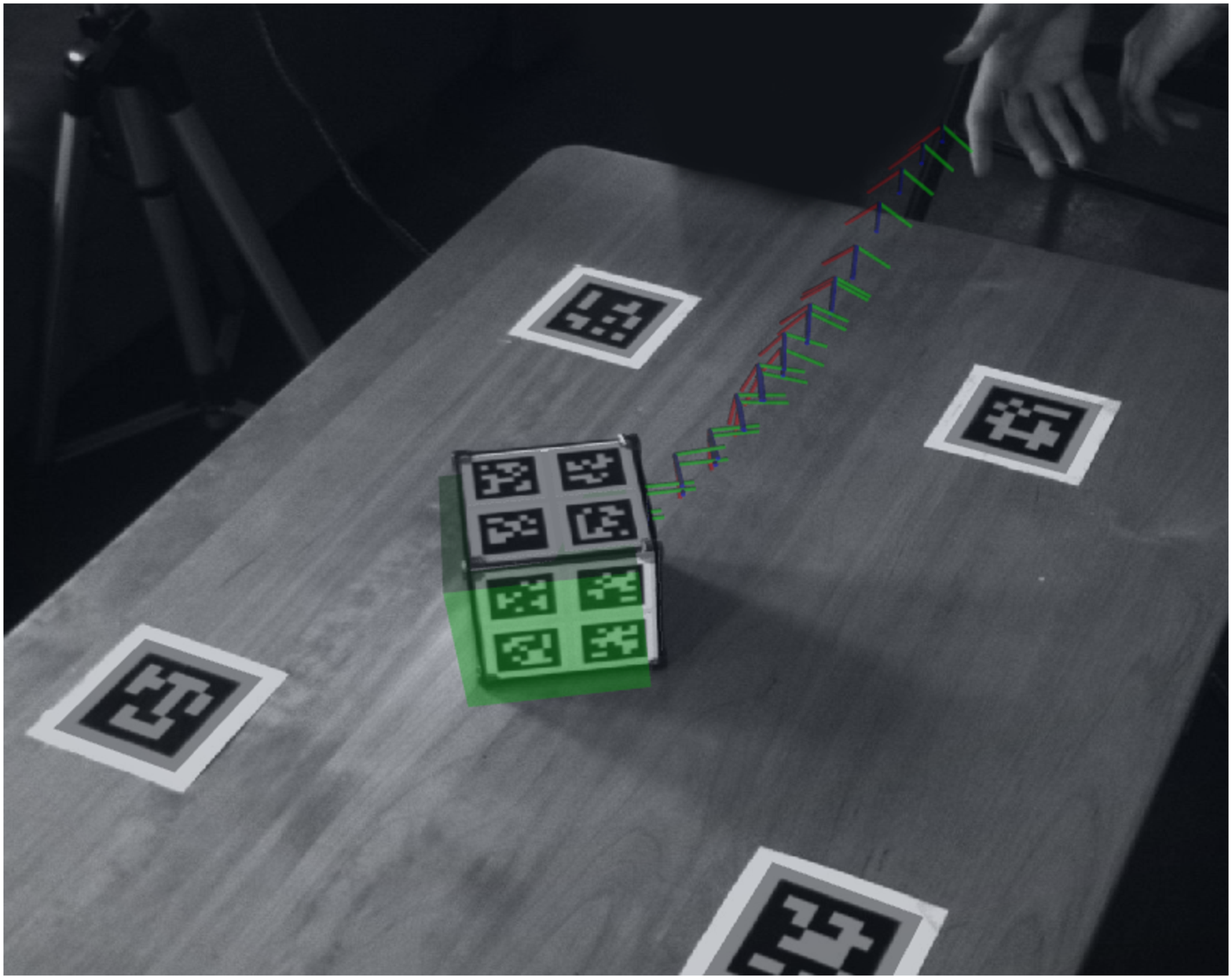}
    \centering
    \caption{\label{fig:traj} A trace of c.o.m. position and block orientation is superimposed on a typical tossing behavior learned by our method, \textit{ContactNets}. In both real and predicted (green) behavior, the block impacts the ground on a corner and tumbles onto its side before coming to rest. }
\end{minipage}
\hspace{0.01\textwidth}
\end{minipage}
\vspace{-.3cm}
\end{figure*}
In order to evaluate our formulation's ability to learn the dynamics of a real system undergoing frictional contact, we conduct experiments on the setup shown in Figure \ref{fig:setup}; the associated source code\footnote{\url{https://github.com/DAIRLab/contact-nets}} is available online. A $10$ \si{\centi\meter} acrylic cube is tossed onto a wooden surface, which produces nearly instantaneous (i.e., sub-timestep) impact behaviors. Cube corners are covered with a thin layer of gelatinous, low-friction material to produce more regular contact behavior. Each side of the cube features four unique AprilTags tracked by three $148$ \si{\hertz} PointGrey cameras using TagSLAM \cite{Pfrommer2019}. Object configurations are represented as $\Configuration = [\vect p ; \; \vect R]$ consisting of the world-frame center of mass position $\vect p$ and rotation matrix $\vect R$, and velocities are represented as $\Velocity = [\dot{\vect p}; \vect \omega]$, where $\vect \omega$ is the cube's body-frame angular velocity. After post-processing the original $750$ tosses, the collected dataset contains $570$ unique, high-quality tosses, whose initial toss directions are then randomly rotated about the world $z$ axis.

\subsection{Comparison and metrics}
We compare the data efficiency of \textit{ContactNets} versus an unstructured baseline for long-term prediction and physical realism. For different dataset sizes, we train the models until loss converges on validation data, then we evaluate the models on a separate test dataset. We evaluate long-term prediction via temporally-averaged absolute error over a model rollout; for a particular ground truth trajectory $(\vect{p}_j^*, \vect R_j^*)_{j\in 1,\dots,N}$ we construct a predicted trajectory $(\vect{\hat p}_j, \vect{\hat R}_j)_{j\in 1,\dots,N}$ using \textit{only} the initial condition $(\vect{p}_1^*, \vect{R}_1^*)$ by recursing through the model's dynamics, and compute our metrics as

\begin{equation}
    e_{pos} = \frac{1}{N} \sum _{j=1}^N \left\lVert \vect{\hat p}_j - \vect{p}^*_j\right\rVert_2\,, \qquad
    e_{rot} = \frac{1}{N} \sum _{j=1}^N |\mathrm{angle}(\vect R_j^*,\vect{\hat R}_j)|\,,
\end{equation}
where $\mathrm{angle}(\vect R^*,\vect{\hat R})$ is the angle of the rotation between $\vect R^*$ and $\vect{\hat R}$. To evaluate physical realism, we additionally examine how much the rollout prediction penetrates the real surface on average:
\begin{equation}
	e_{pen} = \frac{1}{N} \sum _{j=1}^N \mathrm{penetration}(\vect{\hat p}_j,\vect{\hat R}_j)\,.
\end{equation}
\subsection{Models}
\subsubsection{ContactNets Polytope (Ours)}
This model fits a low dimensional representation of the cube, highlighting \textit{ContactNets}' ability to learn simple, discontinuous dynamics from sparse data. A common approximation of polytopes contacting a flat surface is that only the vertices make contact with the ground. For this model, $\Parameters$ contains the body-frame locations of the vertices of the cube and the surface normal. Each $\Gap^{\Parameters}_{n,i}$ transforms the $i$th vertex into the world frame and projects it onto the surface normal. A similar geometric construction is conducted for $\Jt[i]^{\Parameters}$. Detailed equations can be found in Appendix \ref{subsec:Setup}.

Parameters are optimized using the loss \eqref{eq:FinalQPLoss}, and osqpth \cite{Amos2017,Stellato2017} is used to compute its gradient. Forward rollouts are computed using a common contact simulation formulation (\citet{Stewart1996a}). Some additional regularizers are added to prevent simulation artifacts due to the particular behaviors of \cite{Stewart1996a}, and are discussed in detail in Appendix \ref{subsec:Reg}.

\subsubsection{ContactNets Deep (Ours)}
This model extends the polytopic model by adding a DNN to $\Gap_n^{\Parameters}$ and $\Gap_t^{\Parameters}$, e.g.
\begin{equation}
	\Gap_n^{\Parameters}(\Configuration) = \Gap_n^{poly,\Parameters}(\Configuration) + \Gap_n^{DNN,\Parameters}(\Configuration)\,,
\end{equation}
enlarging the model class to include essentially arbitrary object and ground geometries. It is trained in the same fashion as \textit{ContactNets Polytope}.

\subsubsection{End-to-end}
Here we consider a slight modification of a typical unstructured learned dynamics model $\State' = \vect{f}^{\Parameters}(\State, \Input)$. As \textit{ContactNets} only learns to predict the contact impulse, rather than burden the unstructured model with the additional task of identifying continuous dynamics, we instead fit a DNN $\NetForce_{c,DNN}^{\Parameters}(\State, \Input)$ \textit{directly} to the observed contact forces $\NetForce_{c,data}(\State,\Input,\State')$, calculated as in \eqref{eq:ForceFromData}; this model is trained end-to-end on single-step prediction with $L_2$ loss:
\begin{equation}\label{eq:EndToEndForceLoss}
    \Loss_{e2e}(\Parameters,\State,\Input,\State')=\TwoNorm{{\NetForce}_{c,data} - \NetForce_{c,DNN}^{\Parameters}(\State, \Input)}^2\,.
\end{equation}
At test time, motion is predicted using equations \eqref{eq:NewtonsSecondLaw}--\eqref{eq:NetImpulseDecomposition}.

Network architectures and training hyperparameters are discussed in detail in Appendix \ref{subsec:Setup}.

\section{Results}\label{sec:Results}

\begin{figure}[ht]
\centering
\subcaptionbox{\label{subfig:results_pos}}{\includegraphics[width=0.32\textwidth]{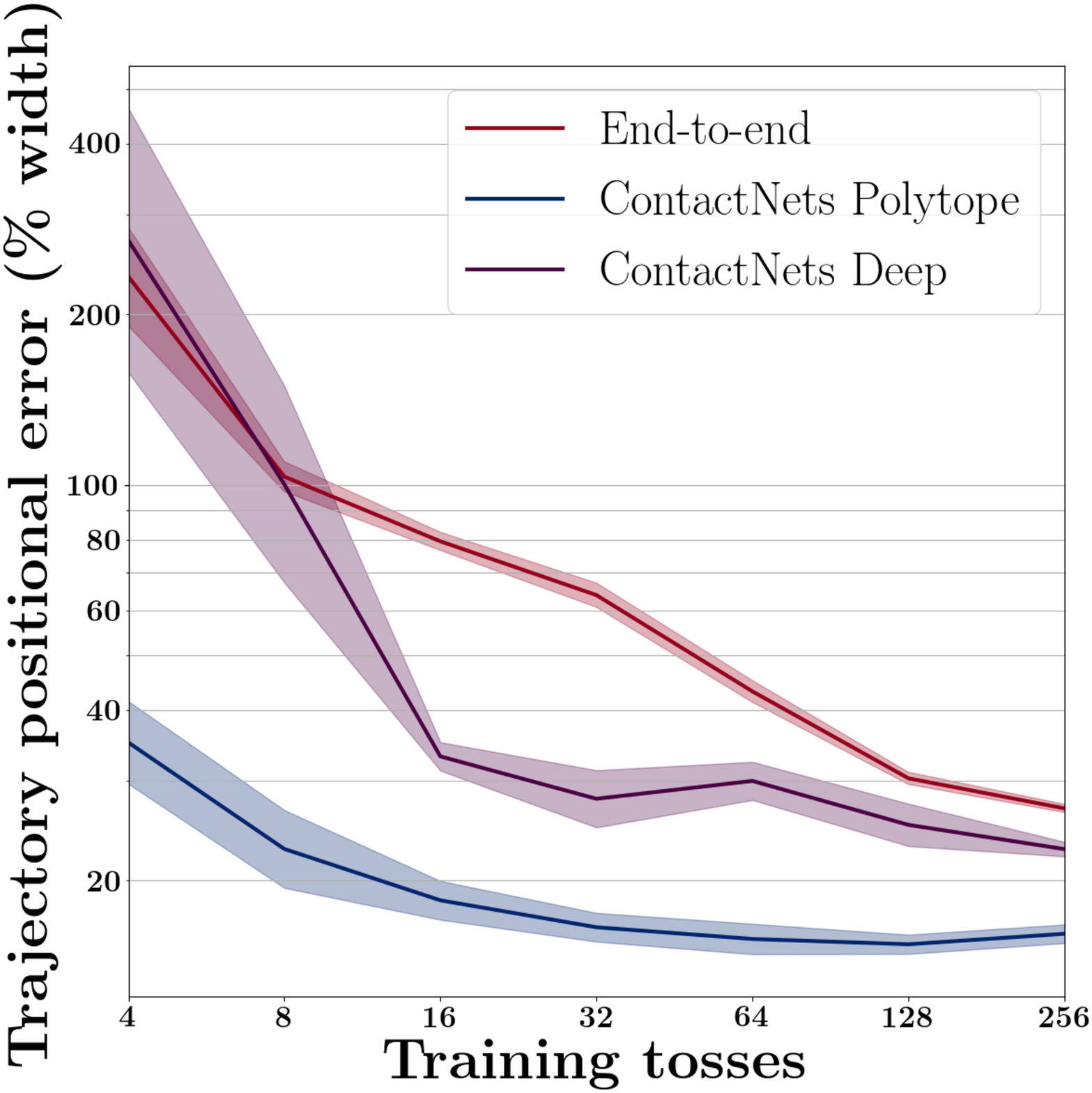}}
\subcaptionbox{\label{subfig:results_rot}}{\includegraphics[width=0.32\textwidth]{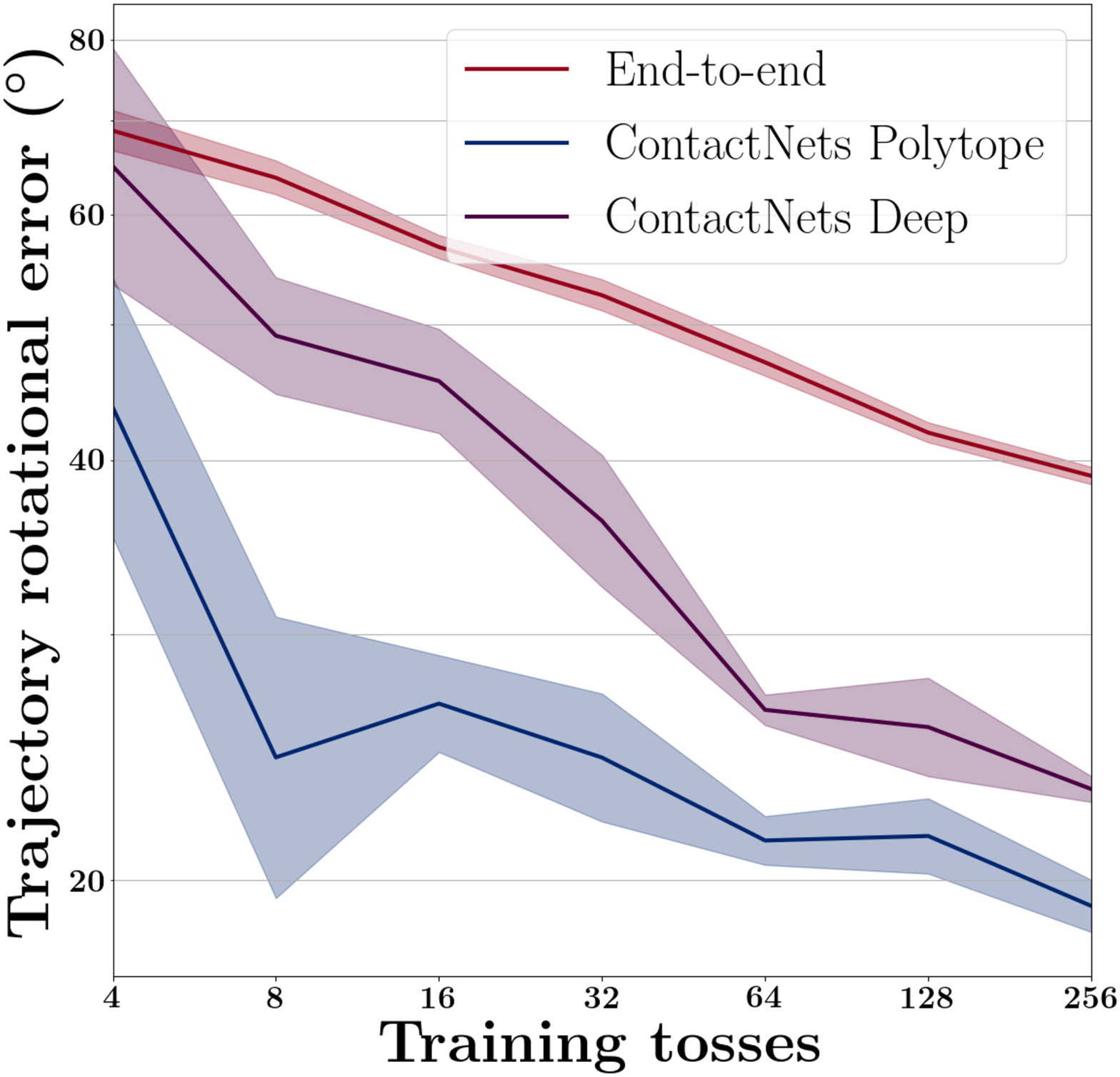}}
\subcaptionbox{\label{subfig:results_pen}}{\includegraphics[width=0.32\textwidth]{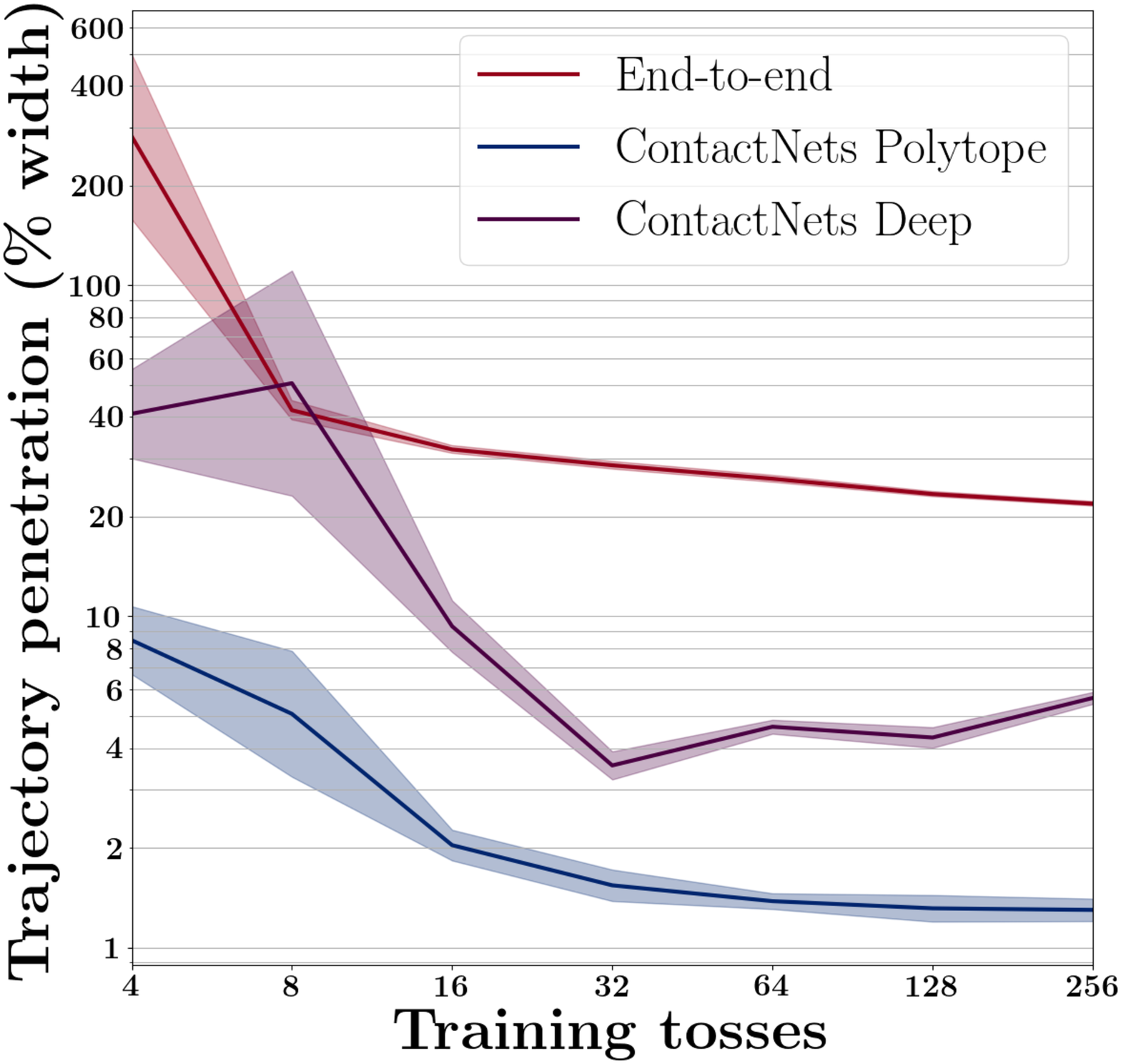}}
\vspace{-.2cm}

\caption{
    (\subref{subfig:results_pos}, \subref{subfig:results_rot}) \textit{End-to-end} model performance is compared to our two parameterizations of contact: \textit{ContactNets Polytope} (low-dimensional) and \textit{ContactNets Deep} (overparameterized). The log-scale graphs show both the mean and $95\%$ confidence interval for a minimum of $4$ samples per point, assuming log-normality of the data. Both \textit{ContactNets} methods achieve at least comparable positional and superior rotational performance to the best $256$-toss \textit{End-to-end} model after just $32$ training tosses. Since the system's geometry is nearly polytopic, \textit{ContactNets Polytope} performs strongly; for more complex interactions we would expect \textit{ContactNets Deep} to have an advantage. (\subref{subfig:results_pen}) Learning representations of inter-body signed distance leads to trajectories which match physical intuition of realistic contact. Despite having no explicit knowledge of ground height or orientation, \textit{ContactNets} model rollouts produce very little penetration, while $\textit{End-to-end}$ methods fail to capture this important behavior.\label{fig:results}}
    
    \vspace{-.3cm}
\end{figure}

We compare the above models in Figure \ref{fig:results}. For a range of data sizes, both \textit{ContactNets} methods outperform the \textit{End-to-end} baseline in positional and rotational accuracy, most strikingly for rotational error with ample training data. \textit{End-to-end} rollouts struggle to capture hard face-to-ground contacts and typically drift rotationally; in contrast, \textit{ContactNets} models are capable of capturing such interactions and are primarily limited by stochastic contact behavior and noisy data.

The penetration metric supports qualitative observerations that \textit{ContactNets} rollouts appear more physically plausible. \textit{End-to-end} models are incapable of producing discontinuous impulses to prevent penetration, and often continue moving after the groud-truth block motion is at rest. This leads to average penetrations of over $20\%$ block width which fail to improve significantly even with more tosses. \textit{ContactNets} rollouts rarely have penetrations of more than $6\%$ block width, with \textit{ContactNets Polytope} averaging at just under $2\%$ block width. Penetrative behaviors exacerbate DNNs' poor ability to extrapolate beyond the training distribution; since these states are non-physical, the training data distribution will never include nearby states, leading learned models to perform poorly. Explicitly encoding complementarity into \textit{ContactNets} eliminates this pathological behavior.

The superior performance of \textit{ContactNets Polytope} compared to \textit{ContactNets Deep} can be attributed to the polytopic geometry of the cube. Further experimentation with curved objects that feature rich, non-isotropic frictional behavior should highlight the more flexible parameterization provided by \textit{ContactNets Deep}. We note that directly parameterizing $\Gap_n$ and $\Gap_t$ as DNNs without a polytopic component proved difficult to train due to the possibility of highly unphysical initializations (i.e., $\Gap_n$ representing a ``ceiling" above the block, instead of a ground below it). We hope to overcome these difficulties with future work incorporating additional initialization and regularization techniques.

\section{Conclusion}
Discontinuous and non-unique impact and stiction underpin essential robotics tasks---thus capturing these phenomena in learned models is crucial for their effective use in the real world.
Our method, \textit{ContactNets}, presents a novel approach to resolving fundamental problems in representing these behaviors with neural networks, and produces realistic dynamics from sparse training data.

The primary limitation of our model is the constrictive nature of its priors: namely that the analytical contact-free dynamics are exact, collisions are inelastic, and objects are rigid.
In future work, we will extend the method to learn continuous forces, and examine models of elastic impact that are consistent with our parameterization, e.g. \citet{Anitescu1997}.
Additionally, as real-time data of object poses is unavailable in some applications, natural extensions could involve embedding our formulation into dynamical models based on visual data. Recent advances in keypoint-based approaches \cite{Manuelli2019} suggest a promising intermediate representation for inferring contact geometry from video.
Further experimentation involving a manipulator interacting with several objects would allow us to evaluate our formulation's ability to capture multi-body contact, and we will verify the quality of our learned models for executing robotic tasks by utilizing them in planning and control algorithms.

\section{Acknowledgements}
This work was supported by the National Science Foundation under Grant No. CMMI-1830218, an NSF Graduate Research Fellowship under Grant No. DGE-1845298, and a Google Faculty Research Award.
We sincerely thank Bernd Pfrommer for assisting with TagSLAM and the camera set-up.

\bibliography{references}
\appendix
\section{Appendix}
\subsection{1D Toy Example}\label{subsec:1DAppendix}
\begin{figure}[h]
\centering
\subcaptionbox{1D System}{\includegraphics[width=0.15\textwidth]{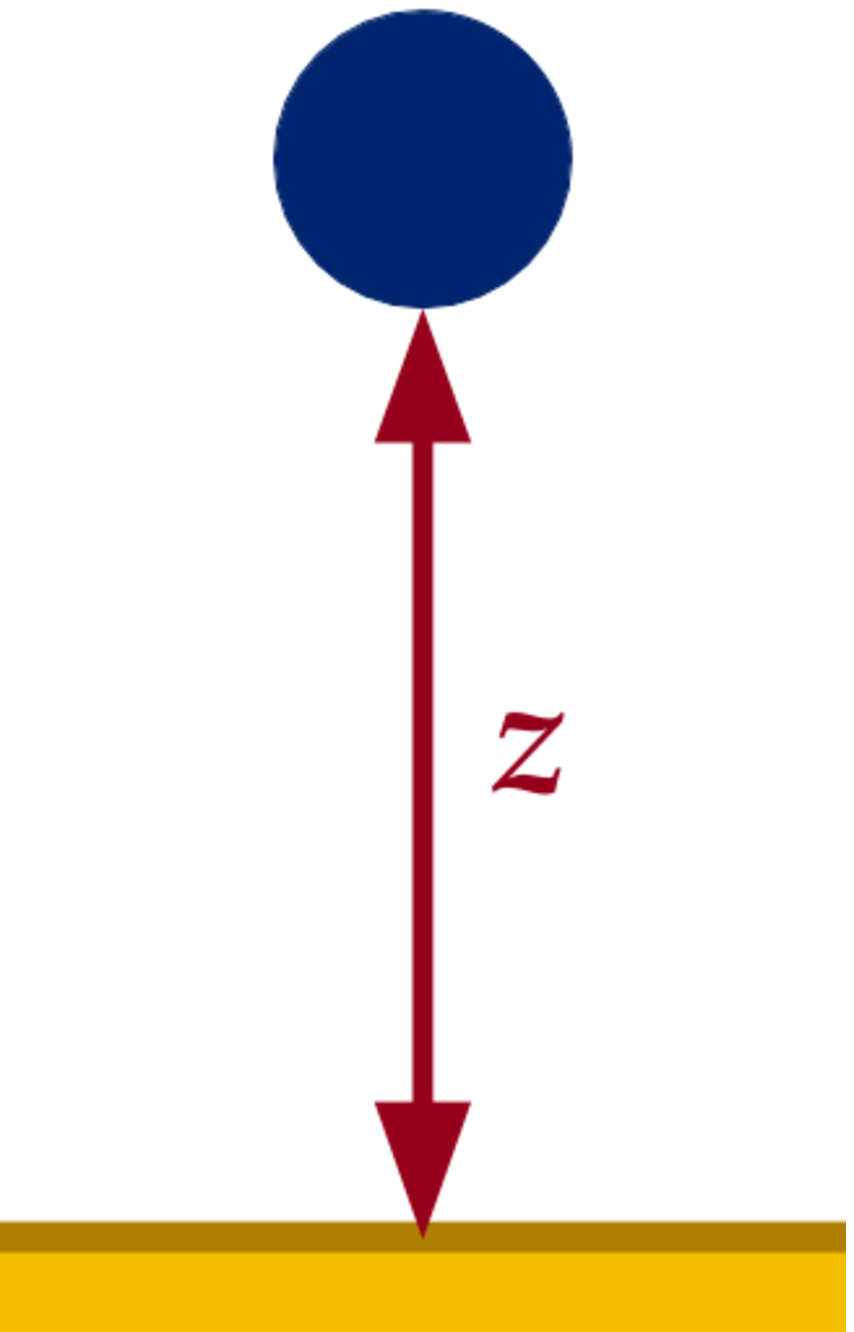}
\vspace{4ex}
}
\subcaptionbox{Model Predictions}{\includegraphics[width=0.54\textwidth]{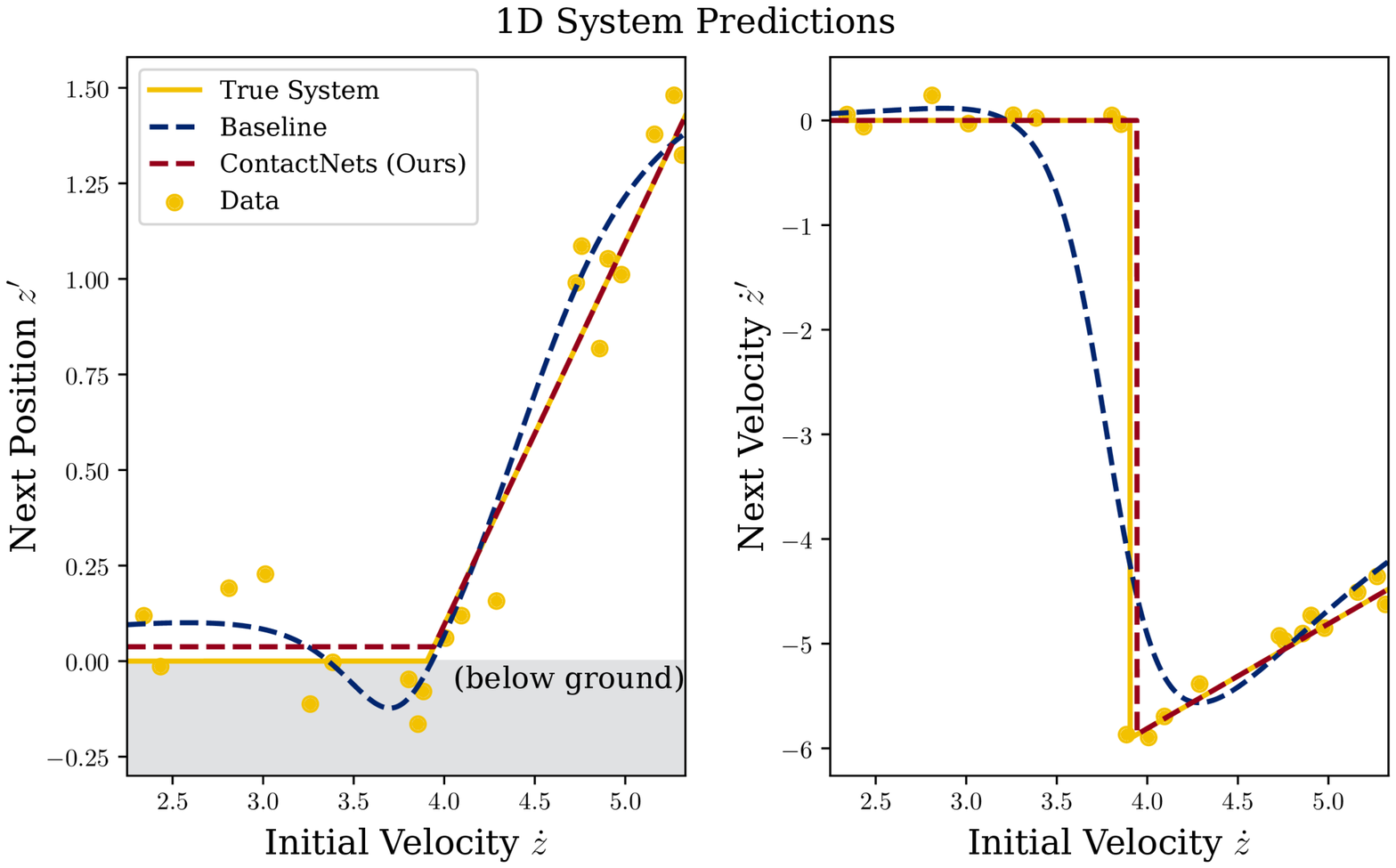}}
\subcaptionbox{Loss Landscape}{\includegraphics[width=0.27\textwidth]{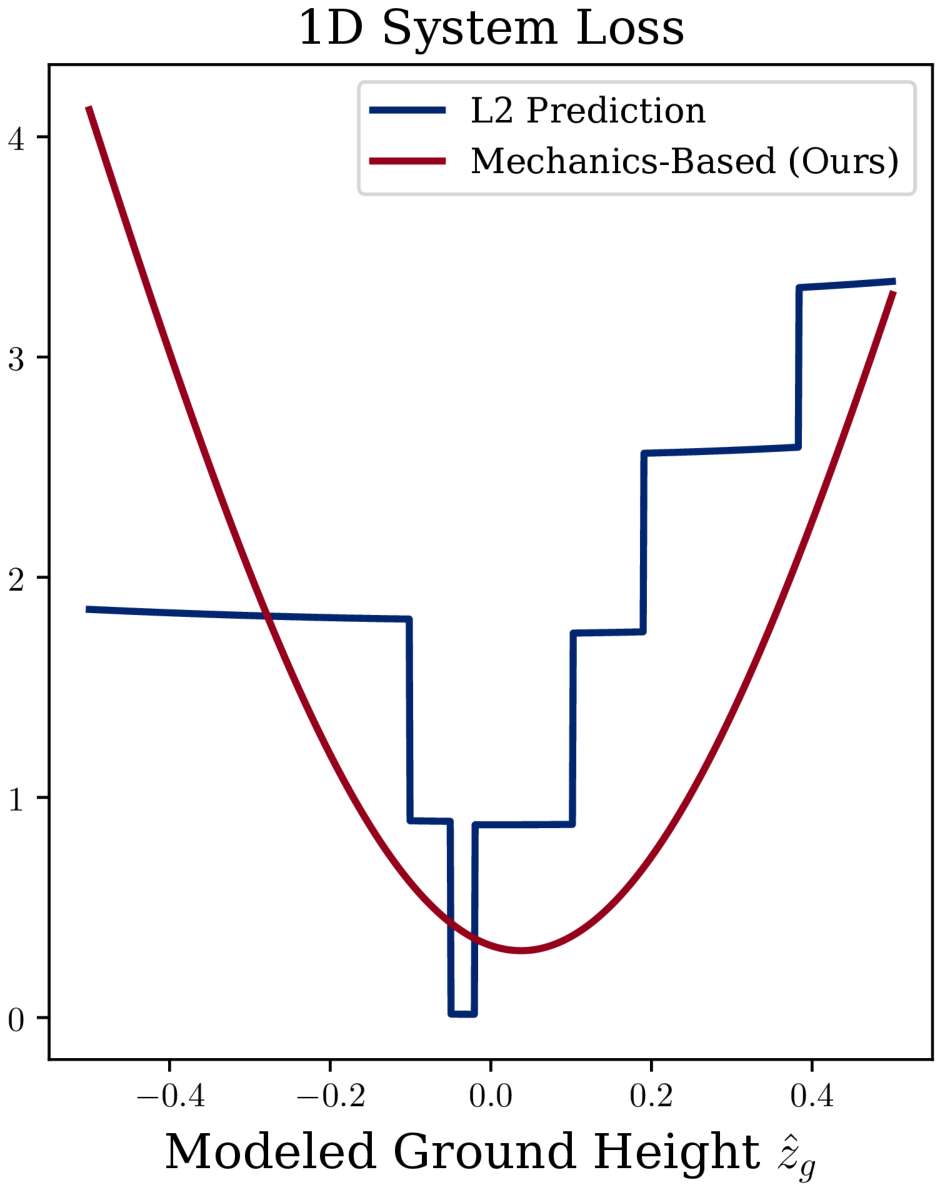}}

\end{figure}
Here, we describe the toy example first shown in Figure \ref{fig:1DExample}, which is reproduced for reference.
\subsubsection{System Dynamics}
In (\subref{subfig:1DPoint}), we display a simple, 1D system with contact: a point mass which makes inelastic impact with the ground at height $z = z_g = 0$. The system has state $\State = [z; \dot z]$ which has freefall motion
\begin{align}
	z'_{free}(z,\dot z) &= z + \dot z \Delta t - \frac{9.81}{2}\Delta t^2\,,\\
	\dot z'_{free}(z,\dot z) &= \dot z - 9.81 \Delta t\,.
\end{align}
For a ground height $z_g$, the next state $\State' = [z'; \dot z]$ either obeys freefall motion, or impacts the ground and comes to rest:
\begin{equation}
	\begin{bmatrix}
		z' \\ \dot z'
	\end{bmatrix} = f_{z_g}(z,\dot z) = \begin{cases}
		\begin{bmatrix}
			z'_{free}(z,\dot z) \\
			z'_{free}(z,\dot z)
		\end{bmatrix} & z'_{free}(z,\dot z) \geq z_g\,, \\
		\begin{bmatrix}
			z_g \\
			0
		\end{bmatrix} & z'_{free}(z,\dot z) < z_g\,.
	\end{cases}\label{eq:1DDynamics}
\end{equation}
In (\subref{subfig:1DPrediction}), we fix the ground height to $z_g = 0$ and the initial position to $z = 1$ and plot $f_{0}(1,\dot z)$ in yellow for $\Delta t = 1$. Note the velocity discontinuity due to impact near $\dot z = 4$.
\subsubsection{Dynamics Learning}
To illustrate the difficulty of fitting a DNN to discontinuity, we consider a simple, unstructured end-to-end dynamics learning setting for the system \eqref{eq:1DDynamics} where we fix $z=0$ and learn the mapping $\dot z \to \State'$. Specifically, we generate training state transitions $(\dot z_i, \State_i ')_{i\in 1,\dots,20}$, by uniformly sampling points from the graph of $f_{0}(1,\dot z)$ in (\subref{subfig:1DPrediction}), and perturbing each element of $\dot z$ and $\State'$ with Gaussian white noise with variance $0.01$. The data is shown in (\subref{subfig:1DPrediction}) as yellow dots.

We train a fully connected DNN $f_{\Parameters}$ to predict $\State_i' \approx f_\theta(\dot z_i)$. The DNN has $2$ hidden layers of width $128$ and $\tanh$ activations, and is trained using Adam with PyTorch default parameters using $L_2$ loss
\begin{equation}
	\mathcal L(\Parameters,\dot z,\State') = \TwoNorm{\State' - f_{\Parameters}(\dot z)}^2\,.
\end{equation}
Training is terminated when the loss converges on a separate, identically distributed validation set. The fully trained network's output is plotted in blue in (\subref{subfig:1DPrediction}). 
The trained DNN is unable to capture the velocity discontinuity well, and predicts significant ground penetration. While stopping early on validation loss prevents the model from overfitting to the noisy training data, the result of this regularization merely produces a smooth regressor, and does not recover important qualitative features of the true system. By a similar notion, any naive regularization that encourages smoothness (e.g. weight decay) will generate similar learned models. Furthermore, an unregularized training process is likely to produce an interpolator of the data, which would exacerbate ground penetration and generate erratic behavior near the discontinuity.

Next, we consider a simple application of \textit{ContactNets} to the 1D system. Without tangential motion along the surface, there are no frictional behaviors in the system; we therefore forego learning related quantities. We consider the simple case of learning an approximation of the ground height $\hat z_g$. As in Section \ref{sec:Approach}, we construct the inter-body signed distance, $\Gap^{\hat z_g}(z) = z - \hat z_g$, and contact impulse estimate $\NetForce_{c,data}(\dot z,\State') = (\dot z' - \dot z) - (-9.81)$. Finally, as there are no frictional behaviors in the system, we construct a simplified version of our mechanics-inspired loss \eqref{eq:FinalQPLoss}:
\begin{equation}
	\Loss(\hat z_g,\dot z,\State') = \min_{\NormalForce \geq 0}\quad \Gap^{\hat z_g}(z')^2\NormalForce^2 + (\NetForce_{c,data} - \NormalForce)^2\,.\label{eq:1DLoss}
\end{equation}

The average of this loss over the data is plotted in (\subref{subfig:1DLoss}) in red for different $\hat z_g$. We learn $\hat z_g$ by minimizing \eqref{eq:1DLoss} using Adam with identical training hyperparameters and termination conditions as the DNN model. After recovering a good estimate for ground height, we can predict the next state as $\hat \State ' = f_{\hat z_g}(1,\dot z)$, shown in red in (\subref{subfig:1DPrediction}). As we embed the key behaviors of contact directly into our model, we both quantitatively and qualitatively outperform the unstructured baseline model. Despite significant noise in the training and validation data, our method produces a ground height $\hat z_g$ which closely approximates the true $z_g$ in the underlying system.

Given that we predict state transitions using $f_{\hat z_g}(1,\dot z)$, it might seem natural employ $L_2$ loss $\TwoNorm{\State' - f_{\hat z_g}(1,\dot z)}^2$, shown in blue in (\subref{subfig:1DLoss}), during training. However, because $f_{\hat z_g}$ is discontinuous in $\hat z_g$, the $L_2$ loss is not differentiable or even continuous, leading to numerical challenges. By contrast, our loss is smooth, allowing higher-order methods like Adam to perform well.

\subsection{Learning setup}\label{subsec:Setup}
\begin{figure}[ht]
\includegraphics[width=0.95\textwidth]{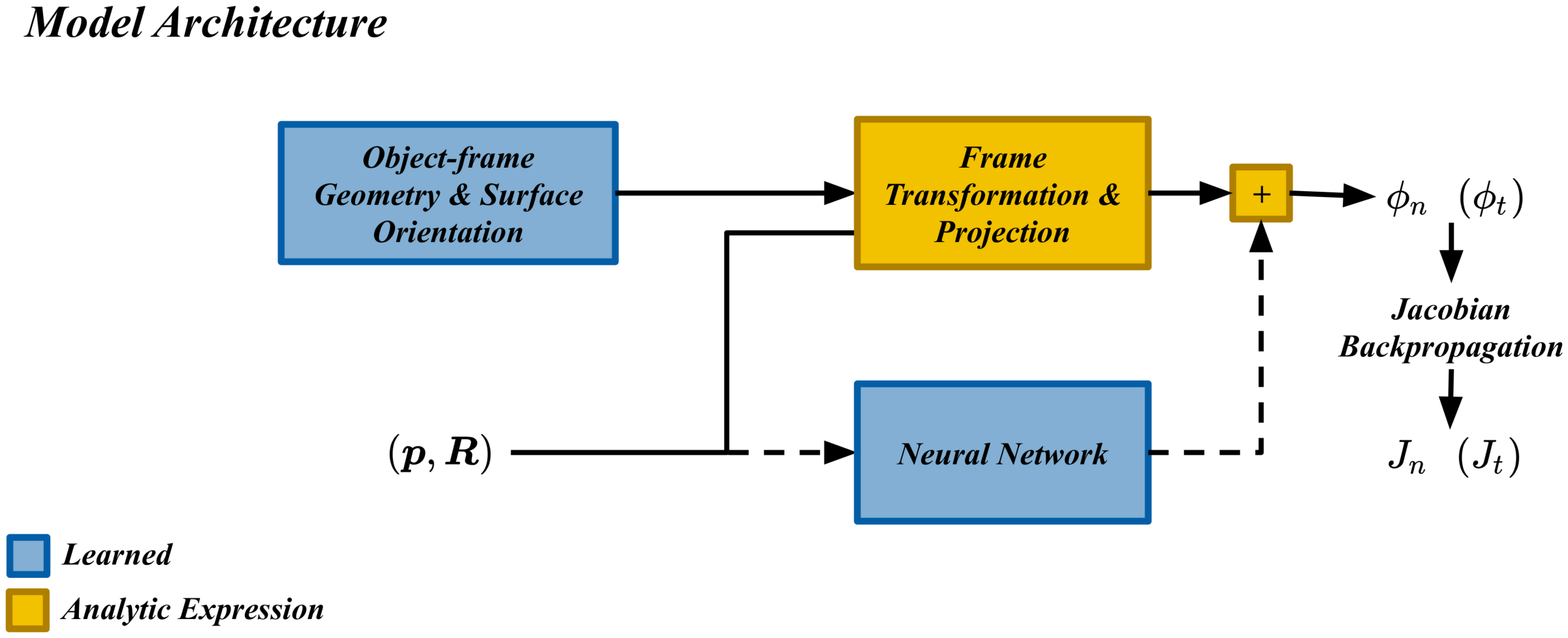}
\centering
\caption{\label{fig:model} Proposed model architectures for learning signed distance and contact-frame Jacobian $\Gap,\J$. \textit{ContactNets Polytope} is represented by the solid lines above, and parameterizes $\Gap_n / \Gap_t$ in terms of the object-frame block geometry and surface orientation. \textit{ContactNets Deep} adds, in parallel, an overparameterized neural network directly mapping from configuration to $\Gap$, as indicated by the dashed lines. Jacobians $\Jn$ and $\Jt$ are computed using Jacobian backpropagation, as described in Appendix \ref{subsec:Setup}.}
\end{figure}

The optimal network structure for \textit{End-to-end} was empirically determined by varying network width and depth, resulting in $4$ hidden layers with $256$ neurons and $\mathrm{ReLU}$ activations. Network inputs are normalized to have zero mean and unit variance. Training is executed using the PyTorch AdamW optimizer with a learning rate of $3 \cdot 10^{-5}$ and weight decay of $10^{-3}$.

The \textit{ContactNets} models are parameterized as depicted in Figure \ref{fig:model}. Object-frame geometry and surface orientation vectors are initialized randomly to their ground-truth values with significant added noise (standard deviation of $40\%$ of their original values).  For \textit{ContactNets Deep}, a separate network is summed in parallel, featuring two hidden layers of $256$ neurons with $\tanh$ activation. Following the addition of regularizers as described in Appendix \ref{subsec:Reg} with coefficients $0.3$, AdamW was used for optimization with a learning rate of $5 \cdot 10^{-4}$ and $0$ weight decay.

For the \textit{ContactNets} methods, we require an additional procedure for computing $\Jn^{\Parameters}(\Configuration)$ from the parameterization of $\Gap_n^{\Parameters}(\Configuration)$ and a given configuration. This is accomplished by first forwards propagating an input $\Configuration$ through the network, keeping note of its value before each operation (activation, weight multiplication, etc.), and then backpropagating a Jacobian matrix using the chain rule. Coupling $\Gap_n^{\Parameters}$ and $\Jn^{\Parameters}$ is critical to ensuring that our learned model produces physically reasonable behavior. We similarly parameterize $\Jt^{\Parameters}$ as the Jacobian of a learned function $\Gap_{t}^{\Parameters}$.

For all models the train-validation-test split is 50-30-20. Each model is trained until its loss fails to improve on the validation set for atleast $12$ epochs (smaller datasets were permitted additional epochs) and is subsequently evaluated on the test dataset in Figure \ref{fig:results}.

\subsection{Learning regularizers}\label{subsec:Reg}
The LCP-based, semi-implicit method of \citet{Stewart1996a} is used to simulate rollouts with the learned $(\Gap_{n,i}^{\Parameters},\J[i]^{\Parameters})$. To prevent unrealistic simulation artifacts, the following regularizers $\mathcal R_1, \mathcal R_2$ were added to the loss \eqref{eq:FinalQPLoss}:
\subsubsection{Normal--tangent perpendicularity} For each contact, we expect that contact-frame forces applied to the body due to normal and frictional contact forces to be orthogonal by definition; hence, we encourage the corresponding elements of the normal and tangential contact Jacobians $\Jn[i] ^{\Parameters}$ and $\Jt[i] ^{\Parameters}$ to be perpendicular by penalizing their normalized dot products:
$$\mathcal R_1 = \sum_i \TwoNorm{\left(\frac{\Jn[i]^{\vect{p}}}{\TwoNorm{\Jn[i]^{\vect{p}}}}\right)
\left(\frac{\Jt[i]^{\vect{p}}}{\TwoNorm{\Jt[i]^{\vect{p}}}}\right)^T}^2\,.$$
 Here, $\Jn[i]^{\vect{p}} = \frac{\partial \Gap_{n,i}^{\Parameters}}{\partial \vect{p}}$ and $\Jt[i]^{\vect{p}} = \frac{\partial \Gap_{t,i}^{\Parameters}}{\partial \vect{p}}$ denote the columns of $\Jn[i]^{\Parameters}$ and $\Jt[i]^{\Parameters}$ that relate to the center of mass position.
\subsubsection{Position Jacobian unit norm}
Regardless of object or table geometry, geometric analysis would imply that $\Jn[i]^{\vect{p}}$ has unit norm. We therefore additionally penalize
 $$\mathcal R_1 = \sum_i \Parentheses{ \TwoNorm{\Jn[i]^{\vect{p}}} - 1}^2\,.$$

\end{document}